\documentclass[12pt,a4paper]{article}

\usepackage{graphicx}

\def\noi{\noindent}
\def\E{{\mathcal E}}
\def\F{{\mathcal F}}
\def\N{{\mathcal N}}
\def\D{{\mathcal D}}
\def\S{{\mathcal S}}

\begin{document}

\date{}

\title{\bf Artificial Neurons with Arbitrarily Complex Internal Structures}

\author{G.A. Kohring\\
C\&C Research Laboratories, NEC Europe Ltd. \\
        Rathausallee 10, D-53757  St. Augustin, Germany
       }

\maketitle

\bigskip
\bigskip

\begin{abstract}
Artificial neurons with arbitrarily complex internal structure are 
introduced.  The neurons can be described in terms of a set of 
internal variables, a set activation functions which describe the time
evolution of these variables and a set of characteristic functions which
control how the neurons interact with one another. 
The information capacity of attractor networks composed of these
generalized neurons is shown to reach the maximum allowed bound.
A simple example taken from the domain of pattern recognition demonstrates 
the increased computational power of these neurons.
Furthermore, a specific class of generalized neurons gives rise to a simple
transformation relating attractor networks of generalized neurons to 
standard three layer feed-forward networks. Given this correspondence, we
conjecture that the maximum information capacity of a three layer 
feed-forward network is 2 bits per weight.
\end{abstract}


\bigskip
\bigskip
\bigskip

\noindent (Accepted for publication in \textit{Neurocomputing}.)

\newpage

\section{Introduction}

The typical artificial neuron used in neural network research today has its 
roots in the McCulloch-Pitts~\cite{MCP43} neuron. It has a 
simple internal structure consisting of a single variable, representing the 
neuron's state, a set of weights representing the
the input connections from other neurons and an 
activation function, which changes the neuron's state.
Typically, the activation function depends upon
a sum of the product of the weights with the state variable of the connecting
neurons and has a sigmoidal shape, although
Gaussian and Mexican Hat functions have also been used. In other words, 
standard artificial neurons implement a simplified version of the
\textit{sum-and-fire} neuron introduced by
Cajal~\cite{GMS83} in the last century.

Contrast this for a moment to the situation in biological systems, where 
the functional relationship between the neuron spiking rate and the membrane
membrane potential is not so simple, depending as it does on a host of
neuron specific parameters~\cite{GMS83}. 
Furthermore, even the notion of \textit{a typical}
neuron is suspect, since mammalian brains consists of many 
different neuron types, many of whose functional role in cognitive 
processing is not well understood.

In spite of these counter examples from biology, the standard neuron
has provided a very powerful framework 
for studying information processing in artificial neural networks.  
Indeed, given the success of current models such as those
of Little-Hopfield \cite{LIT74,HOP82},
Kohonen~\cite{TK83} or Rumelhart, Hinton and Williams~\cite{PDP86},
it might be questioned whether
or not the internal complexity of the neuron plays any significant role
in information processing. In other words, is there any pressing reason to 
go beyond the simple McCulloch-Pitts neuron?

This paper examines this question by considering neurons
of arbitrary internal complexity. Previous researchers have attempted to
study the affects of increasing neuron complexity by adding biologically
relevant parameters, such as a refraction period or time delays, 
to the neuro-dynamics (see e.g. Clark et al., 1985).
The problem with such investigations is that they have so far failed to 
answer the question of whether such parameters are simply an artifact of 
the biological nature of the neuron or whether the parameters are really 
needed for higher-order information processing. To date networks with more
realistic neurons look more biologically plausible, but their processing power
is not better than simpler networks. An additional problem with such studies,
is that as more and
more parameters are added to the neuro-dynamics, software implementations
becomes too slow to
allow one to work with large, realistically sized networks. Although using 
silicon neurons~\cite{CM89} can solve the computational problem, they 
introduce their own set of artifacts which may add or detract from their 
processing power.

The approach taken here differs from earlier work by extending the neuron
while keeping the neuro-dynamics simple and tractable. In doing so, we 
will be able to generalize the notion of the neuron as a processing unit,
thereby moving beyond the biological neuron to include 
a wider variety of information processing units. 
(One has to keep in mind, 
that the ultimate goal of the artificial neural network program is 
not to simply replicate the human brain, but to uncover
the general principles of cognitive processing, so as to perform it more
efficiently than humans are capable of.) 
As a byproduct of this approach, we will demonstrate a formal
correspondence between attractor networks composed of generalized artificial
neurons and the common three layer feed-forward network.

The paper is organized as follows: In the next section,
the concept of the generalized artificial neuron is introduced and its 
usefulness in
attractor neural networks is demonstrated, whereby the 
information capacity of such networks is calculated. 
Section three presents a simple numerical comparison between 
networks of generalized artificial neurons and the conventional multi-state
Hopfield model.
Section four discusses 
various forms that the generalized artificial neuron can take and the 
meaning to be attached to them. 
Section five discusses generalized generalized neurons with interacting 
variables.
The paper ends with a discussion on the
merits of the present approach. Proofs and derivations are relegated to the
appendix.

\section{Generalized Artificial Neurons (GAN) \label{sec:gan_nn}}

Since its introduction in 1943 by McCulloch and Pitts, the 
artificial neuron with a single internal variable (hereafter referred to as
the McCulloch-Pitts neuron)
has been a standard component of artificial neural networks. The neuron's
internal variable may take on only two values, as in the original 
McCulloch and Pitts model, or it may take on a continuum of values.
Although, even where analog or continuous neurons are used, it is 
usually 
a matter of expediency, e.g., learning algorithms such as back-propagation 
\cite{PDP86} require continuous variables even if the 
application only makes use of a two state representation.

Whereas the McCulloch-Pitts neuron presupposes 
that a single variable is sufficient to describe the internal state of 
a neuron, we will generalize this notion by allowing neurons with multiple
internal variables.
In particular, we will describe the internal state of a neuron by $Q$ 
variables.

Just as biological neurons have no knowledge of the internal states of 
other neurons, but only exchange electro-chemical signals (Shepherd, 1983),
a generalized artificial neuron (GAN) should not be allowed knowledge of the 
internal states
of any other GAN. Instead, each GAN has a set of, $C$,
characteristic functions, 
${\bf f}\equiv \{f_i:R^Q\rightarrow R,\ i=1,\dots,C\}$,
which provide
mappings of the internal variables onto the reals.
It is these characteristic functions which are
accessible by other GANs. Even though the characteristic
functions may superficially
resemble the neuron firing rate, no such interpretation need be imposed upon
them.

As in the case of McCulloch-Pitts neurons, the time evolution of the 
internal variables of a GAN are described by a deterministic dynamics.
Here we distinguish between the different dynamics of the $Q$ internal 
variables by
defining $Q$ activation functions, $A_i$. These activation functions may 
depend only upon the values returned by the characteristic functions of the
other neurons.

A GAN, $\N({\bf Q},{\bf f}, {\bf A})$, is thus described by a 
a set of internal variables, ${\bf Q}$, a set of activation functions,
${\bf A}$, and set of characteristic functions, ${\bf f}$. 
Note, for the case of McCulloch-Pitts neuron, there is only a single
internal variable governed by
single activation function taking on one of two 
values: $0$ or $1$, which also doubles as the characteristic function.

Now, to combine these neurons together into a network, we must define a 
network topology. The topology is usually described by a set of numbers, 
$\{W_{ij}\}$ ($i,j=1,\dots,N$), called variously by the names couplings, 
weights, connections or synapses,
which define the edges of a graph having the neurons sitting on the nodes.
(In this paper we will use the term ``weight'' to denote these numbers.)
Obviously, many different network topologies are definable,
each possessing its own properties, therefore,
in order to make some precise statements, let us consider a specific
topology, namely that of a fully connected attractor network 
\cite{LIT74,HOP82}.
Attractor networks form a useful starting point because they are mathematically
tractable and there is a wealth of information already known about them.

\subsection{Attractor Networks \label{sec:att_nn}}

For simplicity, consider the case where each of the $Q$ internal variables
is described by a single bit, then
the most important quantity of interest is the
information capacity per weight, $\E$, defined as:

\begin{eqnarray}
\E &\equiv& 
        \frac{{\rm Number\  of\  bits\  stored}}{{\rm Number\  of\  weights}} 
\label{eq:info_gen}
\end{eqnarray}

\begin{center}
\begin{figure}[htbp]
	\includegraphics[height=20.0cm,keepaspectratio]{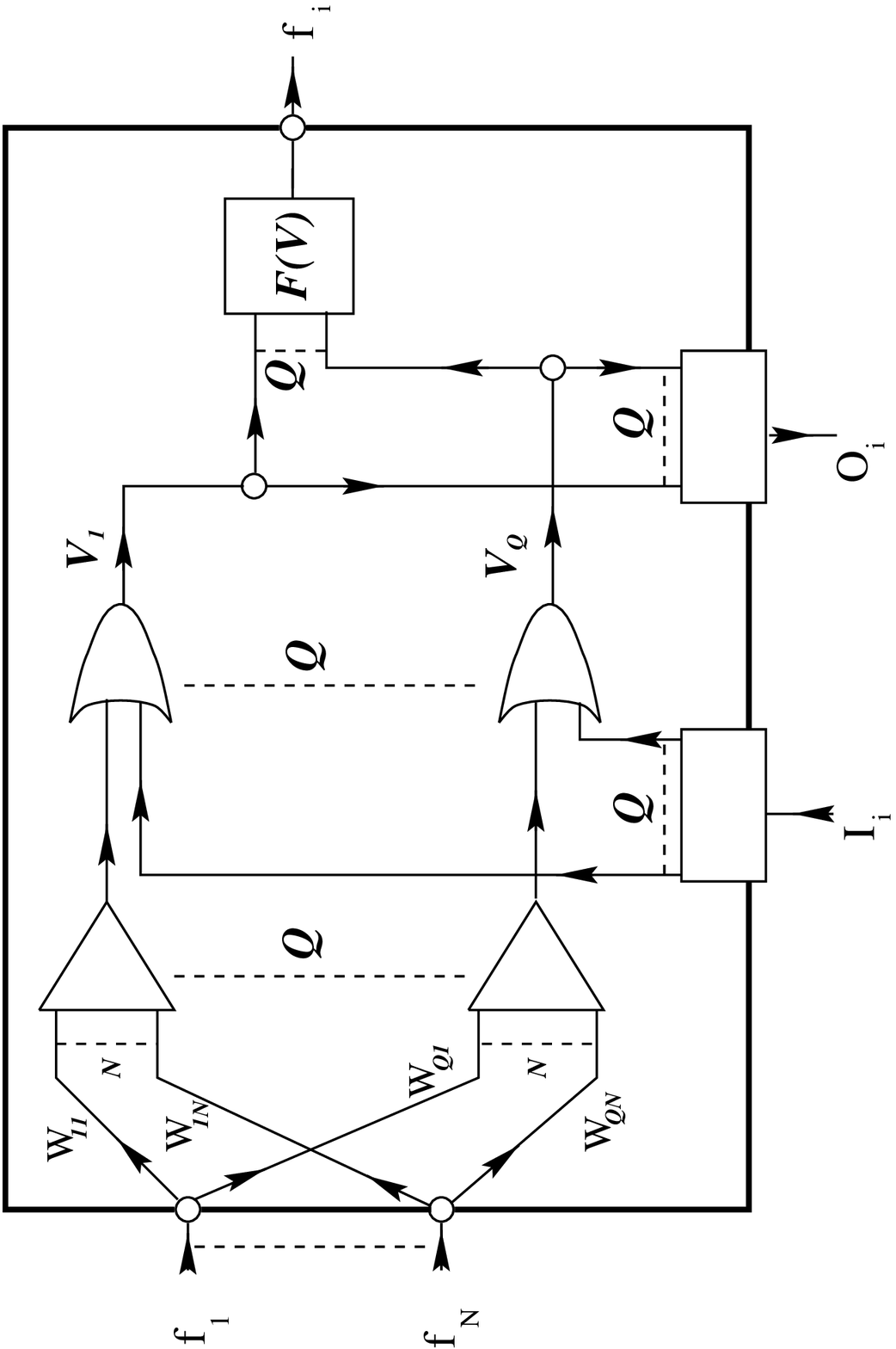}
    \caption{\label{fig:gan} A schematic of a generalized artificial neuron. $f_i$ denotes the value
of the $i$-th neuron's characteristic function, these are the values
communicated to other neurons in the network. $I_i$ and $O_i$ denote input
and output values used for connections external to the network.}
\end{figure}
\end{center}

\noi For a GAN network the number of weights can not simply be the number of
$\{W_{ij}\}$, otherwise it would be difficult for each internal variable to
evolve independently. The simplest extension of the standard topology is to
allow each internal variable to multiply the weights it uses by an independent
factor. Hence, instead of $\{W_{ij}\}$ we effectively have $\{W_{ij}^a\}$, 
where, $a=1,\dots,Q$. A schematic of this type of neuron is given
in Figure~\ref{fig:gan}.  In an attractor network, the goal is to store $P$
patterns such that the network functions as an auto-associative, or
error-correcting memory. The information capacity, $\E$, for these types of
networks is then:

\begin{eqnarray}
\E   &=& \frac{Q PN}{QN^2} \ \ {\rm bpw}, \nonumber \\
 \   &=& \frac{P}{N} \ \ {\rm bpw},
\label{eq:info}
\end{eqnarray}

\noi (bpc $\equiv$ bits per weight) 
As is well known, there is a fundamental limit on the
information capacity for attractor networks, namely $\E \leq 2$ bpw
\cite{TMC68,EG88,GAK90,MKB91}.
This implies, $P \leq 2N$.

Can this limit be reached with a GAN? To answer this question, consider the
case where the activation functions are simply Heaviside functions, $H$:

\begin{eqnarray}
V^a_i(t+1) &=& H\left( \sum_{j\not=i}^NJ_{ij}^aI_N^j(t) \right) \nonumber \\
O^i_N(t+1) &=& F\left( V^1_i(t+1),\dots,V^Q_i(t+1)\right)
\label{eq:fund_dyn}
\end{eqnarray}

\noi where $H(x)=0$ if $x<0$ and $H(x)=1$ if $x \ge 0$. $s^a_i(t+1)$
represents the the $a$-th internal variable of the $i$-th neuron.
The weight to the internal
states of the $i$-th neuron does not violate the principle stated
above, because the $i$-th neuron still has no knowledge of the 
internal states of the other neurons and each neuron is free to adjust
its own internal state as it sees fit.

In appendix A we use Gardner's weight space approach~\cite{EG88} to 
calculate the 
information capacity for a network defined by Eq.~\ref{eq:fund_dyn}, 
where we now take into account the fact that the total number of weights has
increased from $N^2$ to $QN^2$.
Let $\rho$ denote the probability that $s^a = 0$ and $1-\rho$ the
probability that $s^a = 1$, then $\E$ for Eq.~\ref{eq:fund_dyn} becomes:

\begin{equation}
 \E = \frac{-\rho\ln_2\rho - (1-\rho)\ln_2(1-\rho)}
           {1-\rho +\frac{1}{2}(2\rho-1){\rm erfc}(x/\sqrt{2})}\ \ {\rm bpw},
\label{eq:info_new}
\end{equation}

\noi where $x$ is a solution to the following equation:

\begin{equation}
    (2\rho-1)\left[\frac{e^{-x^2/2}}{\sqrt{2\pi}} - 
     \frac{x}{2}{\rm erfc}(x/\sqrt{2})\right] = (1-\rho)x,
\label{eq:aux_eq}
\end{equation}

\noi and ${\rm erfc}(z)$ is the complimentary error function:
${\rm erfc}(z)=(2/\sqrt{\pi})\int_z^{\infty}dy\,e^{-y^2}$.

When $\rho=1/2$, i.e., when $s$ has equal probability of being 
$0$ or $1$, then $x=0$ and the
information capacity reaches its maximum bound of $\E~=~2$~bpw. For highly
correlated patterns, e.g., $\rho\rightarrow 1$, the information capacity
decreases somewhat, $\E\rightarrow 1/(2\ln2)$~bpw, but, more importantly, 
it is still independent of $Q$.

What we have shown is that networks of GANs store
information as efficiently as networks of McCulloch-Pitts neurons. The
difference being, that in the former, each stored pattern contains $NQ$ bits 
of information instead of $N$. Note: we have neglected the number of bits 
needed
to describe the characteristic functions since they are proportional to $QN$,
which for large $N$ is much smaller than the number of weights, $QN^2$.

\section{A Simple Example \label{sec:example}}

Before continuing with our theoretical analysis, let us consider a simple,
concrete example of a GAN network that illustrates their advantages over
conventional neural networks.  Again, we consider an
attractor network composed of GANs.
Each GAN has two internal bit-variables ${\bf Q}=\{s_1,s_2\}$ whose activation
functions are given by Eq.~\ref{eq:fund_dyn} and 
two characteristic 
functions, ${\bf f}=\{g,h\}$. Let $g\equiv q_1\otimes q_2$ and 
$h\equiv q_1+2q_2$. In the neurodynamics defined by Eq.~\ref{eq:fund_dyn}
we will use the function $g$, reserving the function $h$ for communication
outside of the network. (There is no reason why I/O nodes should use the same
characteristic functions as compute nodes.)

The weights will be fixed using a generalized Hebbian 
rule~\cite{DOH49,HOP82}, i.e.,

\begin{equation}
W_{ij}^a = \sum_{\mu=1}^P s_i^{a,\mu}f_j^{\mu}
\label{eq:hebb_coup}
\end{equation}

\noi Since this GAN has 4 distinct internal states,
we can compare the performance of our GAN network  to that 
of a multi-state Hopfield model~\cite{HR90}. Define the neuron 
values in the multi-state Hopfield network as $s\in\{-3,-1,1,3\}$ and 
define thresholds at 
$\{-2,0,2\}$. (For a detailed discussion regarding the simulation of
multi-state Hopfield models see 
the work of Stiefvater and M\"uller \cite{SM92}.)

\begin{center}
\begin{figure}[htbp]
	\includegraphics[height=20.0cm,keepaspectratio]{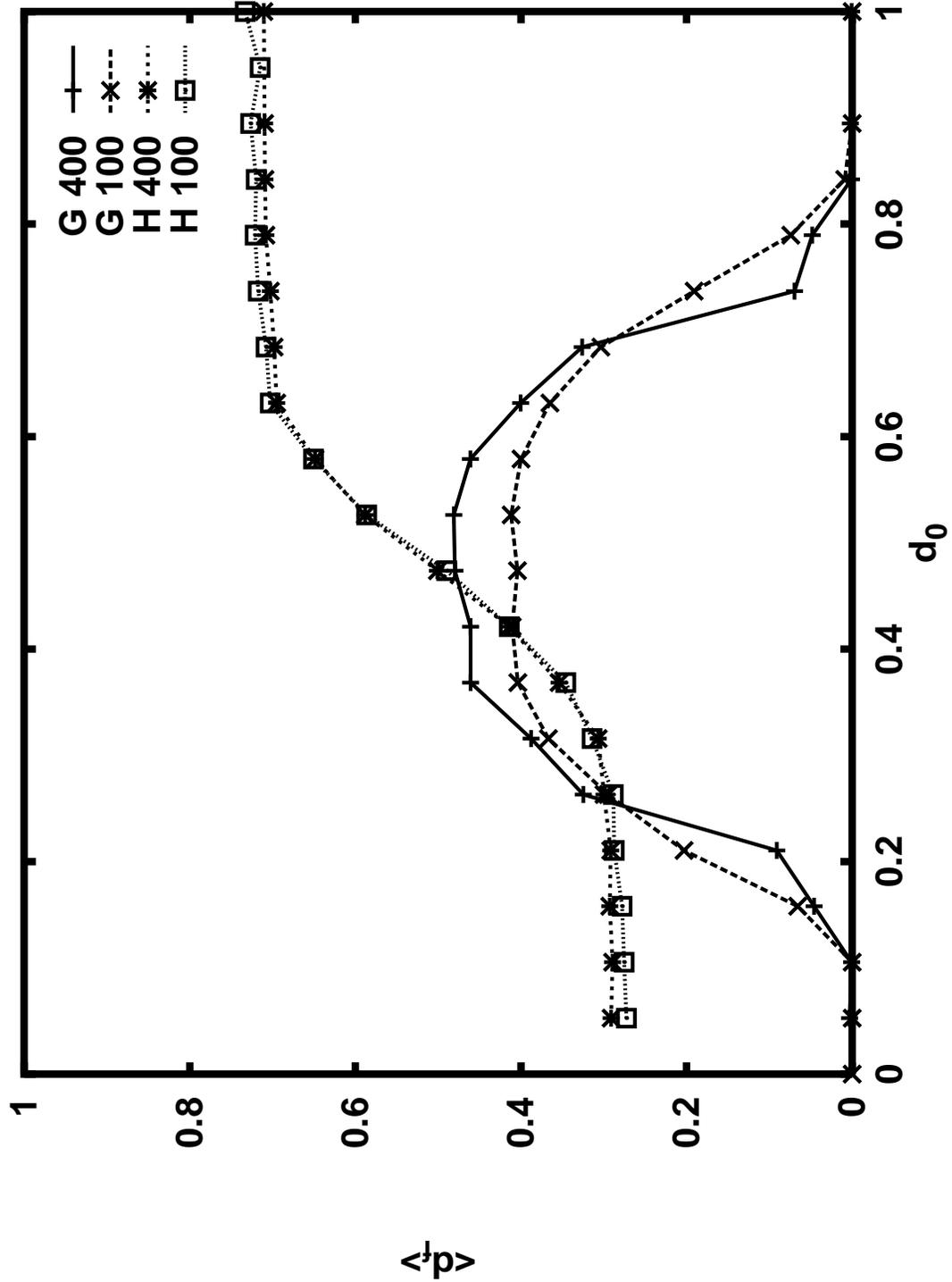}
    \caption{\label{fig:ba} Basins of attraction for a GAN network 
      (lower curves)
      and for a multi-state Hopfield model (upper curve). In both cases
      the number of stored patterns is $P=0.05N$. In each case two
      different system sizes are shown, one with $N=100$ neurons and one
      with $N=400$ neurons.}
\end{figure}
\end{center}

Fig. \ref{fig:ba} depicts the basins of attraction for these two different 
networks, i.e., $d_0$ is the initial distance from a given pattern to 
a randomly chosen starting configuration and $<d_f>$
is the average distance to the same pattern when the network has reached 
a fixed
point. For both network types, random sets of patterns were
used with each set consisting of $P=0.05N$ patterns. The averaging was done 
over all patterns in a given set and over 100 sets of patterns.

There are two immediate differences between the behavior of the multi-state
Hopfield network and the present network: 1) the recall behavior is much
better for the network of GANs, and 2) using the $XOR$ function
as a characteristic function when there are an even number of bit variables,
results in a mapping between a given state and 
its anti-state (i.e., the state in which all bits are reversed), for this
reason the basins of attraction have a hat-like shape instead of the sigmoidal
shape usually seen in the Hopfield model.

This simple example illustrates the difference between networks of
conventional neurons and networks of GANs. Not only is the
retrieval quality improved, but, depending upon the characteristic function,
there is also a qualitative difference in the shape of the basins of 
attraction.

\section{Characteristic Functions}

Until now the definition of the characteristic functions, $\bf f$, has been 
deliberately left open in order to allow us to consider
any set of functions which map the internal variables onto the reals:
${\bf f}\equiv\{f:R^Q\rightarrow R\}$. 
In section \ref{sec:gan_nn} no restrictions on the $f$ were
given, however, an examination of the derivation in appendix A, reveals that 
the characteristic functions do need to satisfy some mild 
conditions before Eq.~\ref{eq:info_new} holds: 

\begin{eqnarray}
1)& \mid\langle f\rangle\mid \ll \sqrt{N},& \nonumber \\
2)& \langle f^2\rangle \ll N,& \textrm{and} \nonumber \\
3)& \langle f^2\rangle -\langle f\rangle^2 \not= 0. &
\label{eq:conditions}
\end{eqnarray}

\noi The first two conditions are
automatically satisfied if $f$ is a so-called squashing function, i.e,
$f:R^Q\rightarrow~[0,1]$.

\subsection{Linear $f$ and Three Layer Feed-Forward Networks \label{sec:ff}}

One of the simplest forms for $f$
is a simple linear combination of the internal variables. Let the internal
variables, $s_i^a(t)$, 
be bounded to the unit interval, i.e., $s_i^a \in [0,1]$, and let $J_i^a$
denote the coefficients associated with the $i$-th neuron's $a$-th internal
variable, then $f$ becomes:

\begin{equation}
f_i(t) = \sum_{a=1}^Q J_i^a s_i^a(t).
\label{eq:weighted_sum}
\end{equation}

\noi Provided, $\mid \sum_{a=1}^QJ_i^a \mid \ll \sqrt{N}$, and provided 
not all $J_i^a$
are zero, the three conditions in Eq.~\ref{eq:conditions} will be satisfied. 
Since the internal variables are bounded to the unit interval, let their
respective activation functions be any sigmoidal function, $\S$.
Then we can substitute $\S$ into
Eq.~\ref{eq:weighted_sum} in order to obtain a
time evolution equation solely in terms of the characteristic functions:

\begin{equation}
f_i(t) = \sum_{a=1}^Q J_i^a{\S}
                 \left( \sum_{j\not=i}^NW_{ij}^af_j(t-1) \right).
\label{eq:feed_forward}
\end{equation}

\noi Formally, this equation is, for a given $i$, equivalent to that of a 
three layer neural network with $N-1$ linear neurons on the input layer, 
$Q$ sigmoidal neurons in the hidden layer and one linear neuron on the output
layer. From the work of Leshno et al.\footnote{Leshno et al.'s proof is the
most general in a series of such proofs. For earlier, more restrictive
results see e.g., \cite{KF89,HSW89,HK91}},  we know that three
layer networks of this form are sufficient to approximate any continuous
function $F:R^{N-1}\rightarrow R$ to any degree of accuracy provided $Q$ is 
large
enough. Leshno et al.'s result applied to Eq.~\ref{eq:feed_forward} shows 
that at each time step, a network of $N$ GANs is capable 
of approximating any continuous function $F:R^N\rightarrow R^N$ 
to any degree of accuracy.

In section \ref{sec:att_nn} the information capacity of a GAN attractor network
was shown to be given by the solution of eqs. \ref{eq:info_new} and 
\ref{eq:aux_eq}. Given the formal correspondence demonstrated above,
the information capacity of a conventional three layer neural network must
be governed by the same set of equations. Hence, the maximum 
information capacity in a conventional three layer network is limited 
to 2 bits per weight.

\subsection{Correlation and Grandmother functions}

A special case of the linear weighted sum discussed above is
presented by the correlation function: 

\begin{equation}
f_i(t) = \frac{1}{Q}\sum_{a=1}^Q t_i^a s_i^a(t),
\label{eq:overlap}
\end{equation}

\noi where the $\{t_i^a\}$ represent a specific configuration of the internal
states of $\N(Q,f)$. With this form for $f$, the GANs can
represent symbols using the following interpretation for $f$: 
as $f\rightarrow 1$, the symbol is present, 
and as $f\rightarrow 0$ the symbol is not present. 
Intermediate values represent the symbols partial 
presence as in fuzzy logic approaches.  In this scheme, a symbol is 
represented locally, but
the information about its presence in a particular pattern is distributed.
Unlike other representational schemes, by increasing the number of
internal states, a symbol can be represented by itself. Consider, for example, 
a pattern recognition system. If $Q$ is large enough, one could represent
the symbol for a tree by using the neuron firing pattern for a tree. In this
way, the symbol representing a pattern is the pattern itself.
 
Another example for $f$ in the same vein as Eq.~\ref{eq:overlap} is given by:

\begin{equation}
f_i(t) = \delta_{\{s_i^a(t)\},\{t_i^a\}},
\label{eq:grandma}
\end{equation}

\noi where, $\delta_{x,y}$ is the Kronecker delta function: $\delta_{x,y} =1$
iff $x=y$. This equation states that $f$ is one when the value of all
internal variables are equal to their values in some predefined 
configuration. A GAN of this type represents what is
sometimes called a grandmother cell.

\subsection{Other Forms of $f$}

Obviously, there are an infinite number of functions one could use for $f$, 
some of
which can take us beyond conventional neurons and networks, to a more 
general view of computation in neural network like settings. Return for a
moment to the example discussed in section \ref{sec:example}:

\begin{equation}
f_i(t) = \bigotimes_{a=1}^Q s_i^a(t).
\label{eq:xor}
\end{equation}

\noi This simply implements the parity function over all
internal variables. Its easy to see that $\langle f\rangle = 1/2$ and
$\langle f^2\rangle -\langle f\rangle^2=1/4$, hence, this form of $f$
fulfills all the necessary conditions. Using the $XOR$ function as a
characteristic function for a GAN trivially solves Minsky and
Papert's objection to neural networks \cite{MP69} at the
expense of using a more complicated neuron. 

Of course Eq.~\ref{eq:xor} can be generalized to represent 
any Boolean function. In fact, each $f_i$ could be a different Boolean
function, in which case the network would resemble the Kauffman model for 
genomic systems \cite{SAK92}, a model whose chaotic behavior and 
self-organizational properties have been well studied.

\section{Neurons with Interacting Variables}

So far we have considered only the case where the internal variables of the
GAN are coupled to the characteristic function of other 
neurons and not to each other, however,  in principle, there is no reason 
why the internal variables should not interact. For simplicity consider once
again the case of an attractor network. The easiest method for 
including the internal variables in the dynamics 
is to expand Eq.~\ref{eq:fund_dyn} by adding a new set of weights, 
denoted by, $\{L_{i}^{ab}\}$, which couple the internal variables to each
other:

\begin{equation}
s_i^a(t+1) = H\left(\sum_{j\not= i}^NW_{ij}^af_j +
          \sum_{b\not= a}^QL_{i}^{ab}s_i^{b} \right).
\label{eq:intern_dyn}
\end{equation}

\noi Using the same technique we use in section \ref{sec:att_nn}, we can
determine the new information capacity for attractor networks (see appendix
A):

\begin{equation}
\E = \E_0
\frac{
     \left[1+\lambda\sqrt{\frac{\rho(1-\rho)}
                    {\langle \phi^2\rangle -\langle\phi\rangle^2}}\,\right]^2
     }{ (1+\lambda)\left(1+\lambda\frac{\rho(1-\rho)}
                    {\langle \phi^2\rangle -\langle\phi\rangle^2}\right)
     },
\label{eq:intern_info}
\end{equation}

\noi where, $\E_0$ is given by Eq.~\ref{eq:info_new}, $\lambda\equiv Q/N$ and
$\langle\phi\rangle$ is the average value of the characteristic
function at the fixed points. From this equation we see that 
if the 
fluctuations in the characteristic functions are equal to the fluctuations in the
internal variables, then $\E = \E_0$, otherwise, $\E$ is always less than
$\E_0$.

\section{Summary and Discussion}

In summary, we have introduced the concept of the generalized 
artificial neuron (GAN), 
$\N({\bf Q},{\bf f}, {\bf A})$, where ${\bf Q}$ is a set of internal 
variables, $\bf f$ is a set characteristic functions acting on those 
variables and $\bf A$ is a set of activation functions describing the
dynamical evolution of those same variables.
We then showed that the information capacity of
attractor networks composed of such neurons reaches the maximum allowed
value of 2 bits per weight. If we use a linear characteristic function \`a~la
Eq.~\ref{eq:weighted_sum}, then we find a relationship between three layer
feed forward networks and attractor networks of GANs. This
relationship tells us that attractor networks of GANs can 
evaluate an arbitrary
function of the form $F:R^N\rightarrow R^N$ at each time step. Hence, their
computational power is significantly greater than that of attractor 
networks with two state neurons. 

As an example of the increased computation power of the GAN, we presented a
simple attractor network composed of four state neurons. The present network
significantly out performed a comparable multi-state Hopfield model. Not only
were the quantitative retrieval properties better, but the qualitative
features of the basins of attraction were also fundamentally different. It is
this promise of obtaining qualitative improvements over standard models that
most sets the GAN approach apart from previous work.

In section \ref{sec:att_nn}, the upper limit on the information 
capacity of an attractor
network composed of GANs was shown to be 2 bits per weight, while, in
section \ref{sec:ff} we demonstrated a formal correspondence between these
networks and conventional three layer feed-forward networks. Evidently, 
the information capacity results apply to the more conventional feed-forward
network as well.

The network model presented here bears some resemblance to models
involving hidden (or latent) variables (see e.g., \cite{HS86}),
however, there is one important difference: namely, the hidden variables
in other models are only hidden in the sense that they are isolated from 
the network's inputs and outputs; but they are
not isolated from each other, they are allowed full participation 
in the dynamics, including direct interactions with one another. 
In our model, the internal neural variables interact
only indirectly via the neurons' characteristic functions.

Very recently, Gelenbe and Fourneau \cite{GF99} proposed a related 
approach they
call the ``Multiple Class Random Neural Network Model''. Their model also
includes neurons with multiple internal variables, however, 
they do not distinguish between activation and characteristic functions,
furthermore, they restrict the form of the activation function to be a 
stochastic variation of the usual
sum-and-fire rule, hence, their model is not as general as the one 
presented here.

In conclusion, the approach advocated here can be used to exceed the
limitations imposed by the McCulloch-Pitts neuron. By increasing the 
internal complexity we have been able to increase the computational power
of the neuron, while at the same time avoiding any unnecessary increase 
in the complexity of the neuro dynamics, hence, there should
be no intrinsic limitations to implementing our generalized artificial neurons.

\newpage

\bibliographystyle{acm}
\bibliography{nn}

\newpage
\noindent{\large\bf Figures}

\newpage

\appendix

\section{Derivation of the Information Capacity}

For simplicity consider a homogeneous network of 
$N$ GANs, where the $Q$ internal variables of each neuron are
simply bit-variables. In addition, we will consider the general case
of interacting bits.  Given $P$ patterns, with $\phi_i^{\mu}$ representing
the characteristic functions and $\sigma_i^{a\mu}$ the internal bit-variables,
then by equation eqs. \ref{eq:fund_dyn} and \ref{eq:intern_dyn}, we see 
that these patterns will be fixed points if:

\begin{equation}
(2\sigma_i^{a\mu}-1)\left( \sum_{j\not= i}^NW_{ij}^a\phi_i^{\mu} +
          \sum_{b=1}^QL_{i}^{ab}\sigma_i^{b\mu} \right) > 0.
\label{eq:stability}
\end{equation}

\noi In fact, the more positive the left hand side is, the more stable the
fixed points. Using this equation we can write the 
total volume of weight space available to the network for storing $P$ 
patterns as:

\begin{equation}
V = \prod_{i,a}V_i^a, \nonumber 
\end{equation}
\noi where,
\begin{eqnarray}
V_i^a=\frac{1}{Z_i^a}\int \prod_{j}dW_{ij}^a\, \prod_bdL_i^{ab}
\ \ \delta\left(\sum_{j\not= i}^N(W_{ij}^a)^2-N\right)
\delta\left(\sum_{b\not= a}^Q(L_{i}^{ab})^2-Q\right)
\ \times 
\nonumber \\
\quad \prod_{\mu}\, H\left[ 
(2\sigma_i^{a\mu}-1)
          \left(\frac{1}{\sqrt{N}}\sum_{j\not= i}^NW_{ij}^a\phi_j^{\mu} +
          \frac{1}{\sqrt{N}}\sum_{b\not= a}^QL_{i}^{ab}\sigma_i^{b\mu} -
\theta_i^a\right) -\kappa \right],
\quad
\label{eq:Volume}\\
\noalign{\noi {\rm and}} \nonumber \\
Z_i^a=\int \prod_{j}dW_{ij}^a \, \prod_bdL_i^{ab}\ 
\delta\left(\sum_{j\not= i}^N(W_{ij}^a)^2-N\right)
\delta\left(\sum_{b\not= a}^Q(L_{i}^{ab})^2-Q\right). \qquad
\end{eqnarray}

\noi where $\kappa$ is a constant whose purpose is the make the left hand
side of Eq.~\ref{eq:stability} as large as possible. 
(Note, although we have introduce a threshold parameter, 
$\theta_i^a$, we will show that thresholds do not affect the results.)

The basic idea behind the weight space approach is that the subvolume, 
$V_i^a$, will vanish for all values of $P$ greater than some critical value, 
$P_c$. In order to find the \textit{average} value of $P_c$, we need to 
average Eq.~\ref{eq:Volume} over all configurations of $\sigma_i^{a\mu}$.
Unfortunately, the $\sigma_i^{a\mu}$ represent a quenched average, 
which means that we have to average the intensive quantities derivable
from $V$ instead of averaging over $V$ directly. The simplest 
intensive such quantity is:

\begin{eqnarray}
\F&=&\lim_{N\rightarrow \infty} 
\langle \ln V_i^a \rangle_{\sigma_i^{a\mu}}, \nonumber \\
&=&\lim_{{N\rightarrow \infty}\atop{n\rightarrow 0}}
\frac{\left\langle \left(V_i^a\right)^n \right\rangle_{\sigma_i^{a\mu}} - 1}{n}.
\end{eqnarray}

\noi The technique for performing the averages in the limit $n\rightarrow 0$ 
is known as the replica method \cite{EA75}.

By introducing integral representations for the Heaviside functions
\newline 
(\ $H(z-\kappa)=\int_\kappa^\infty \, dx\int_{-\infty}^\infty\, dy\,\exp(iyx)$
\ ) 
we can perform the averages over the $\sigma_i^{a\mu}$:

\begin{eqnarray}
\langle V_i^a \rangle_{\sigma_i^{a\mu}} &=&
\sum_{\sigma_j^{b\mu}} \frac{1}{Z_i^{a}}\int\prod_{jA}dW_{ij}^{aA}
      \int_\kappa^{\infty}\prod_{\mu A}dx_\mu^A\int_{-\infty}^\infty
       \prod_{\mu A}dy_\mu^A \times \nonumber \\
&\ & \exp\Biggl\{i\sum_{A=1\atop\mu=1}^{n,P}
         y_{\mu}^A\biggl[ x_{\mu}^A-(2\sigma_i^{a\mu}-1)
\bigl(\frac{1}{\sqrt{N}}\sum_{j\not= i}^NW_{ij}^{aA}\phi_j^{\mu} + \nonumber \\
&\ & \qquad \qquad \qquad
\frac{1}{\sqrt{N}}\sum_{b\not= a}^QL_i^{abA}\sigma_i^{b\mu} -
\theta_i^a\bigr)\biggr]\Biggr\} \times \nonumber \\
&\ & \prod_{A=1}^n\delta\left(\sum_{j\not= i}^N(W_{ij}^{aA})^2-N\right)
\delta\left(\sum_{b\not= a}^Q(L_{i}^{ab})^2-Q\right).
\label{eq:rep_avg}
\end{eqnarray}

\noi First sum over the $\sigma_j^{b\mu}$ where $j\not= i$:

\begin{eqnarray}
&\ &\sum_{\sigma_j^{b\mu}}\exp\left\{-i\sum_{A=1\atop\mu=1}^{n,P}
         y_{\mu}^A(2\sigma_i^{a\mu}-1)
\left(\frac{1}{\sqrt{N}}\sum_{j\not= i}^NW_{ij}^{aA}\phi_j^{\mu}
\right)\right\} = \nonumber \\
&\ & \prod_{j,\mu} \sum_{\sigma_i^{a\mu}}
\exp\left\{-i\frac{(2\sigma_i^{a\mu}-1)}{\sqrt{N}}
\sum_Ay_{\mu}^AW_{ij}^{aA}\phi_j^{\mu}\right\} \approx \nonumber \\
&\ &\prod\limits_{j,\mu} 
\left[1 - i\frac{(2\sigma_i^{a\mu}-1)\langle \phi\rangle}{\sqrt{N}}
\sum_{A}y_{\mu}^AW_{ij}^{aA} - 
       \frac{\langle \phi^2\rangle}{2N}
\sum_{AB}y_{\mu}^Ay_{\mu}^BW_{ij}^{aA}W_{ij}^{aB}\right] \approx \nonumber \\
&\ &\exp\left\{- i\frac{(2\sigma_i^{a\mu}-1)\langle \phi\rangle}{\sqrt{N}}
\sum_{A\mu}y_{\mu}^A\sum_jW_{ij}^{aA} -
\frac{\langle \phi^2\rangle -\langle \phi\rangle^2}{2N}\sum_{AB}\sum_{\mu}
y_{\mu}^Ay_{\mu}^B\sum_jW_{ij}^{aA}W_{ij}^{aB}\right\}, \nonumber \\ &\ &
\label{eq:exp}
\end{eqnarray}

\noi now sum over the $\sigma_j^{b\mu}$ where $j=i$ but $b\not= a$:

\begin{eqnarray}
&\ &\sum_{\sigma_j^{b\mu}}\exp\left\{-i\sum_{A=1\atop\mu=1}^{n,P}
         y_{\mu}^A(2\sigma_i^{a\mu}-1)
\left(\frac{1}{\sqrt{N}}\sum_{b\not= a}^QL_{i}^{abA}\sigma_i^{b\mu}
\right)\right\} \approx \nonumber \\
&\ &\exp\left\{- i\frac{(2\sigma_i^{a\mu}-1)(1-\rho)}{\sqrt{N}}
\sum_{A\mu}y_{\mu}^A\sum_bL_{i}^{abA} -
\frac{\rho(1-\rho)}{2N}\sum_{AB}\sum_{\mu}
y_{\mu}^Ay_{\mu}^B\sum_bL_{i}^{abA}L_{i}^{abB}\right\}, \nonumber \\ &\ &
\label{eq:exp2}
\end{eqnarray}

\noi where we have use $\rho$ as the probability that $\sigma=0$, 
$\langle \phi\rangle \equiv \sum_{\sigma} \phi(\sigma)$ and
$\langle \phi^2\rangle \equiv \sum_{\sigma} \phi(\sigma)\phi(\sigma)$.
If we insert Eq.~\ref{eq:exp} into \ref{eq:rep_avg} and define the
following quantities: $q^{AB}~=~(1/N)\sum_j W_{ij}^{aA}W_{ij}^{aB}$ and 
$r^{AB}~=~(1/Q)\sum_b L_{i}^{abA}L_{i}^{abB}$ for all
$A<B$ and $M^{aA}_{i}=(1/\sqrt{N})\sum_jW_{ij}^{aA}$ and
$T^{aA}_{i}=(1/\sqrt{Q})\sum_bL_{i}^{abA}$ for all $A$,
then Eq.~\ref{eq:rep_avg} can be rewritten as:

\begin{eqnarray}
\langle V_i^a \rangle_{\sigma_i^{a\mu}} \propto
\int\prod_{A}dz^A\, dM^A\, dE^A\, dU^A\, dT^A\, dC^A\, 
\prod_{A<B}\,q^{AB}F^{AB}r^{AB}H^{AB}\: \ e^{NG}, \nonumber \\
\label{eq:steep_int}
\end{eqnarray}

\noi where,

\begin{eqnarray}
G&\equiv& \alpha G_1(q,M,T)+G_2(F,z,E)+\lambda G_2(U,H,C)+
i\sum_{A<B}F^{AB}q^{AB}+ \nonumber \\
&\ & i\lambda\sum_{A<B}H^{AB}r^{AB} +
     \frac{i}{2}\sum_Az^A + \frac{i\lambda}{2}\sum_AU^A + 
       O(1/\sqrt{N}). 
\end{eqnarray}

\noi $\alpha \equiv P/N$ and we have introduced another parameter:
$\lambda\equiv Q/N$. The functions $G_1$ and $G_2$ are defined as:

\begin{eqnarray}
G_1 &\equiv& \frac{1}{P}\ln 
         \Biggl\langle \int_{\kappa}^{\infty}\prod_{\mu A}dx_\mu^A
           \int_{-\infty}^\infty \prod_{\mu A}dy_\mu^A \exp \biggl\{
            i\sum_{A\mu}y_{\mu}^A+ \nonumber \\
&\ & \qquad \qquad
i\sum_{A\mu}y_{\mu}^A(2\sigma_i^{a\mu}-1)
     \left(\theta^a - \langle \phi\rangle M^A - 
\sqrt{\lambda}(1-\rho)T^A\right) \nonumber \\
&\ & \qquad \qquad
       -\frac{\langle \phi^2\rangle -\langle \phi\rangle^2 +
\lambda\rho(1-\rho)}{2}
 \sum_{\mu A}(y_\mu^A)^2 - \nonumber \\
&\ & \qquad \qquad
 \sum_{A<B}\sum_\mu y_\mu^Ay_\mu^B\left[q^{AB}
(\langle \phi^2\rangle -\langle \phi\rangle^2)
+r^{AB}\lambda\rho(1-\rho)\right]
\biggr\}\Biggr\rangle_\sigma 
\nonumber \\ 
&=&
          \ln \Biggl\langle \int_{\kappa}^{\infty}\prod_{A}dx^A
           \int_{-\infty}^\infty \prod_{A}dy^A \exp \biggl\{
            i\sum_{A}y^A+ \nonumber \\
&\ & \qquad \qquad
i\sum_{A}y^A(2\sigma-1)
        \left(\theta - \langle \phi\rangle M^A - 
   \sqrt{\lambda}(1-\rho)T^A\right) \nonumber \\
&\ & \qquad \qquad
       -\frac{\langle \phi^2\rangle -\langle \phi\rangle^2 +
\lambda\rho(1-\rho)}{2}
 \sum_{A}(y^A)^2 -\nonumber \\
&\ & 
  \sum_{A<B}y^Ay^B
\left[ q^{AB}(\langle \phi^2\rangle -\langle \phi\rangle^2) +
r^{AB}\lambda\rho(1-\rho)\right]
\biggr\}\Biggr\rangle_\sigma,
\end{eqnarray}

\noi and

\begin{eqnarray}
G_2(x,y,s) &\equiv& \frac{1}{N}\ln 
         \Biggl[ \int_{-\infty}^{\infty}\prod_{jA}dW_{ij}^{aA}
          \exp \biggl\{-\frac{i}{2}\sum_{A}y^A\sum_j(W_{ij}^{aA})^2-i\sum_{A<B}
           x^{AB}\sum_jW_{ij}^{aA}W_{ij}^{aB} \nonumber \\
    &\ & \qquad \qquad -i\sum_As^A\sum_jW_{ij}^{aA}
           \biggr\} \Biggr] \nonumber \\
&=& \ln
         \Biggl[ \int_{-\infty}^{\infty}\prod_{A}dW^{A}
          \exp \biggl\{-\frac{i}{2}\sum_{A}y^A(W^A)^2-i\sum_{A<B}
           x^{AB}W^AW^B -i\sum_As^AW^A \biggr\} \Biggr]. \nonumber \\
&\ &
\end{eqnarray}

\noi The so-called replica symmetric solution is found by taking 
$q^{AB}~\equiv~q$, $r^{AB}~\equiv~r$, $F^{AB} \equiv F$ and $H^{AB}\equiv H$
for all $A<B$, and setting
$z^A\equiv z$, $U^A\equiv U$, $E^A\equiv E$, $C^A\equiv C$,
$M^A\equiv M$ and $T^A\equiv T$, for all $A$. In terms of
replica symmetric variables, $G_2$ has the form:

\begin{eqnarray}
G_2(x,y,s) &\approx&  -\frac{n}{2}\ln(iy-ix) -\frac{1}{2}\frac{nx}{y-x}-
       \frac{ns^2}{iy-ix}+ O(n^2), \\ &\ & 
\end{eqnarray}

\noi while $G_1$ can be reduced to:

\begin{eqnarray}
G_1 &\approx&  n \int_{-\infty}^{\infty}\D s \; \Biggl\{
\rho \ln I_{-} + (1-\rho) \ln I_+ \Biggr\} + O(n^2),\\
\noalign{\noi where,} \nonumber \\
I_\pm &=& \frac{1}{2}
{\rm erfc}\left(\frac{\kappa\pm v+\sqrt{
q(\langle \phi^2\rangle -\langle\phi\rangle^2) + r\lambda\rho(1-\rho)}\; s}
{\sqrt{2\left[(1-q)\left(\langle \phi^2\rangle -\langle \phi\rangle^2\right)
         +(1-r)\lambda\rho(1-\rho)\right]}}
\right), \nonumber \\
\end{eqnarray}

\noi and we have set $\D s\equiv e^{-s^2/2}/\sqrt{2\pi}$, 
$v\equiv \theta - \langle \phi\rangle M - \sqrt{\lambda}(1-\rho)T$.
$\rm erfc(z)$ is the complimentary error function:
${\rm erfc}(z)\equiv(2/\sqrt{\pi})\int_z^\infty dy\;e^{-y^2}$. Since the
integrand of Eq.~\ref{eq:steep_int} grows exponentially with $N$, we can 
evaluate the integral using steepest descent techniques. The saddle point 
equations which need to be satisfied are:

\begin{eqnarray}
\frac{\partial G}{\partial E} = 0, \quad \frac{\partial G}{\partial C}=0,
\quad
\frac{\partial G}{\partial z} = 0, \quad \frac{\partial G}{\partial U}=0,
\quad
\frac{\partial G}{\partial F}=0, \quad \frac{\partial G}{\partial H}=0, 
\label{eq:sat1} \\
\qquad \frac{\partial G}{\partial q}=0
\qquad {\rm and} \qquad \frac{\partial G}{\partial r}=0.
\qquad \qquad \qquad \qquad \qquad \qquad
\label{eq:sattle}
\end{eqnarray}

\noi Solving this set of equations yields a system of three equations which 
define $q$, $r$ and $v$ in terms of 
$\alpha$ and $\lambda$. A little reflection reveals that when
$\alpha=P/N$ approaches its critical value, $\alpha_c=P_c/N$, 
then $q\rightarrow 1$ and $r\rightarrow 1$, hence, this limit will
yields the critical information capacity. From Eq.~\ref{eq:sattle} the
following relationship between $q$ and $r$ as they both approach 1 can
be deduced:

\begin{equation}
1-r \approx (1-q)\sqrt{\frac{\langle \phi^2\rangle -\langle\phi\rangle^2}
                      {\rho(1-\rho)}}.
\end{equation}

\noi We can now write the information capacity per weight as:

\begin{eqnarray}
\E &=& \left[-\rho\ln_2\rho - (1-\rho)\ln_2(1-\rho)\right]\:
\frac{QPN}{QN^2+NQ^2}
\nonumber \\
&=& \left[-\rho\ln_2\rho - (1-\rho)\ln_2(1-\rho)\right]\:
\frac{\alpha_c}{1 + \lambda},
\end{eqnarray}

\noi with:

\begin{eqnarray}
\alpha_c^{-1} = \Biggl(
\rho\Biggl\{(K-V)\frac{e^{-(K-V)^2/2}}{\sqrt{2\pi}} + 
\frac{1}{2}\left[1+(K-V)^2 \right]
    {\rm erfc}\left(\frac{-K+V}{\sqrt{2}}\right)\Biggr\} + \qquad\nonumber \\
(1-\rho)
\Biggl\{(K+V)\frac{e^{-(K+V)^2/2}}{\sqrt{2\pi}}+ 
\frac{1}{2}\left[1+(K+V)^2 \right] 
{\rm erfc}\left(\frac{-K-V}{\sqrt{2}}\right)\Biggr\} \Biggr)\times
\nonumber \\
\left[1+\lambda\frac{\rho(1-\rho)}
                    {\langle \phi^2\rangle -\langle\phi\rangle^2}\right]/
     \left[1+\lambda\sqrt{\frac{\rho(1-\rho)}
                    {\langle \phi^2\rangle -\langle\phi\rangle^2}}\,\right]^2,
\qquad \qquad
\qquad \qquad
\label{eq:gen_alpha} 
\end{eqnarray} 

\noi where V is implicitly defined through:

\begin{eqnarray} 
\rho\Biggl\{\frac{e^{-(K-V)^2/2}}{\sqrt{2\pi}}+\frac{K-V}{2}
    {\rm erfc}\biggl(\frac{-K+V}{\sqrt{2}}\biggr) \Biggr\} = 
\qquad \qquad \qquad \qquad \qquad
\nonumber \\
(1-\rho)\Biggl\{\frac{e^{-(K+V)^2/2}}{\sqrt{2\pi}} + \frac{K+V}{2}
    {\rm erfc}\left(\frac{-K-V}{\sqrt{2}}\right)\Biggr\}, 
\qquad \qquad 
\nonumber \\
\label{eq:gen_aux}
\end{eqnarray}

\noi and $K\equiv \kappa/\left[\langle \phi^2\rangle -\langle
\phi\rangle^2 + \rho(1-\rho)\right]$.  Note: For a given $\rho$, the 
maximum value of
$\E$ occurs when $K=0$. By setting $K$ and $\lambda$ equal to zero, 
one recovers Eq.~\ref{eq:info_new} and \ref{eq:aux_eq} in the text.
(Its also interesting to note, that since 
$V=\left[\theta - \langle \phi\rangle M - \sqrt{\lambda}(1-\rho)T\right]/
\left[\langle \phi^2\rangle -\langle \phi\rangle^2 + \rho(1-\rho)\right]$,
$M$ and $T$, which represent the average values of the inter and intra neuron
weights respectively, are not uniquely determined, rather solving 
Eq.~\ref{eq:gen_aux} for $V$ only fixes the difference between $T$ and $M$.
Furthermore, the threshold $\theta$ can be easily absorbed
into either $M$ or $T$ provided either $\langle \phi\rangle\not= 0$ or 
$\rho\not= 1$.)

We arrived at equations \ref{eq:gen_alpha} and \ref{eq:gen_aux} using the
saddle point conditions of Eq.~\ref{eq:sat1} and \ref{eq:sattle}. 
As the reader can readily verify, 
these saddle point equations are also locally stable. 
Furthermore, since the volume of the space of allowable weights is connected
and tends to zero as $q,r\rightarrow 1$,
the locally stable solution we have found must be the \textit{unique} 
solution \cite{EG88},
Therefore, in this case, the replica symmetric solution is also the 
exact solution.

\end{document}